\documentclass[letterpaper, 10 pt, conference]{ieeeconf}  

\IEEEoverridecommandlockouts                              

\usepackage{graphics} 
\usepackage{epsfig} 
\usepackage{mathptmx} 
\usepackage{times} 
\usepackage{amsmath} 
\usepackage{amssymb}  
\usepackage{color}
\usepackage[bottom]{footmisc}
\usepackage{booktabs}
\usepackage{tabularx}
\usepackage{hyperref} 
\usepackage{makecell}
\usepackage{multirow}  
\usepackage{cite}
\usepackage{subcaption}
\usepackage{balance}
\usepackage[normalem]{ulem}

\usepackage[para,online,flushleft]{threeparttable}

\begin{document}

\title{\LARGE \bf
Robust Autonomous Vehicle Pursuit without Expert Steering Labels
}

\author{Jiaxin Pan*$^{1}$, Changyao Zhou*$^{1}$, Mariia Gladkova$^{1,4}$, Qadeer Khan$^{1,2}$ and Daniel Cremers$^{1,2,3,4}$
\thanks{* These authors contributed equally.}
\thanks{$^{1}$ Authors affiliated with the Technical University of Munich, TUM}
\thanks{$^{2}$ Authors affiliated with Munich Center for Machine Learning, MCML}
\thanks{$^{3}$ Author affiliated with the University of Oxford.}
\thanks{$^{4}$ Author affiliated with the Munich Data Science Institute.}
}

\maketitle
\thispagestyle{empty}
\pagestyle{empty}

\begin{abstract}

In this work, we present a learning method for both lateral and longitudinal motion control of an ego-vehicle for the task of vehicle pursuit. The car being controlled does not have a pre-defined route, rather it reactively adapts to follow a target vehicle while maintaining a safety distance. To train our model, we do not rely on steering labels recorded from an expert driver, but effectively leverage a classical controller as an offline label generation tool. In addition, we account for the errors in the predicted control values, which can lead to a loss of tracking and catastrophic crashes of the controlled vehicle. To this end, we propose an effective  data augmentation approach, which allows to train a network that is capable of handling different views of the target vehicle. During the pursuit, the target vehicle is firstly localized using a Convolutional Neural Network. The network takes a single RGB image along with cars' velocities and estimates target vehicle's pose with respect to the ego-vehicle. This information is then fed to a Multi-Layer Perceptron, which regresses the control commands for the ego-vehicle, namely throttle and steering angle. We extensively validate our approach using the CARLA simulator on a wide range of terrains. Our method demonstrates real-time performance, robustness to different scenarios including unseen trajectories and high route completion. Project page containing code and multimedia can be publicly accessed here: \href{https://changyaozhou.github.io/Autonomous-Vehicle-Pursuit/}{https://changyaozhou.github.io/Autonomous-Vehicle-Pursuit/}.

\end{abstract}


\section{Introduction}
\label{sec:introduction}

The latest technological advances in the field of autonomous driving have sparked a growing interest in developing more efficient and robust transportation solutions. In this regard, autonomous multi-vehicle platooning and convoying have appeared as a prominent research direction, which promotes various advantages over single vehicle operation such as improved traffic flow, reduced fuel consumption and improved overall transportation efficiency~\cite{li2022review, turri2017model, li2017platoon, zhuang2020robust}.

In this work, we focus on following a chosen vehicle and, therefore, refer to it as vehicle pursuit. This task can be defined as an imitation of the driving behavior of a specific target vehicle without overtaking or violating any traffic rules. In this case, the target vehicle guides the ego-vehicle across a wide spectrum of scenarios and, thus, enables autonomous operation of the ego-vehicle without human intervention. This way, vehicle automation allows to eliminate human errors and ensures precise trajectory control. Moreover, credibility of such methods can only be ascertained via deployment in actual scenarios. Nonetheless, this may be too risky to perform using hardware implementations as it involves direct interaction with the driving environment.

Autonomous driving simulators provide a viable alternative for the training and validation of new approaches for urban driving. Recent open-source platforms, such as~\cite{Carla}, offer multi-sensor high-fidelity data to safely test and evaluate driving algorithms. In addition, recording various driving situations and rare corner cases is made feasible due to flexible and easy-to-control simulation. This has inspired us to choose this platform for the design and deployment of our method for the task of vehicle pursuit.

\begin{figure*}[!h]
\centering
\includegraphics[width=0.8\textwidth]{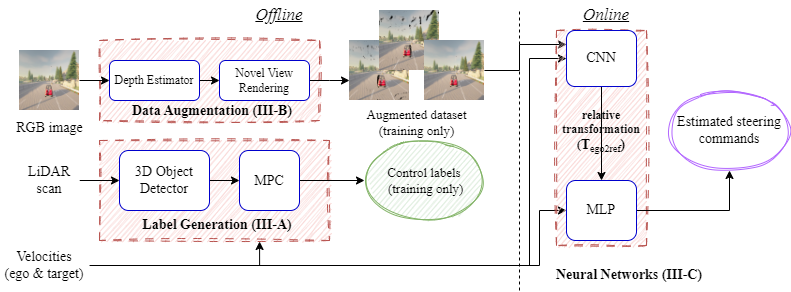}

\caption{\label{overview} \textbf{Overview of proposed framework:} Our proposed framework consists of preparatory offline steps (left) \& online deployment (right), which are respectively done during training \& testing of our system. \textit{Left:} Firstly, we perform label generation using Model Predictive Control - MPC (Sec.\ref{sec:label_generation}). MPC takes location \& orientation of the target with respect to the ego-vehicle estimated by a LiDAR-based 3D detector. Next, data augmentation is done by utilizing estimated dense depth maps for novel view synthesis (Sec. \ref{subsec:data_augmentation}).  \textit{Right:} Our learning method is trained to predict accurate lateral \& longitudinal control values by leveraging obtained labels, augmented image dataset and velocities of both ego \& target vehicles (Sec. \ref{subsec:nn}). At test time, our approach only needs RGB image sequence \& velocities as input for ego-vehicle control.}
\vspace{-7mm}
\end{figure*}

Specifically, we offer a deep learning method for estimating both the lateral and longitudinal steering commands (also referred to as control values) of an ego-vehicle, which pursues a desired target vehicle. Our network relies on a stream of images capturing the target vehicle and the velocities of both cars. The target vehicle is controlled by a human, whereas the ego vehicle is autonomous and controlled by our method. We handle a wide range of scenarios, such as sharp turns, roundabouts, as well as sudden velocity changes of the target car without any human expert supervision and labeling. The latter is achieved as we utilize model predictive control (MPC) for the generation of training labels, including throttle and steering angle. In comparison to other classical controllers, MPC has superior performance as shown by~\cite{9276986,JRC8800}. 

It is important to mention that MPC is only applied offline to extract control labels for training and validation of the neural network and, thus, it is not required online during test time. As we follow a dynamic object, we deploy existing 3D object detectors to localize the target. We make our approach more robust to small inaccuracies in predicted steering commands by incorporating novel views into the training set, which are efficiently generated with the same sensor setup using dense depth maps and an image rendering approach. We demonstrate the full pipeline in Fig.~\ref{overview}.

Our main contributions can be summarised as follows:

\begin{itemize}
\item {We propose a novel framework that enables autonomous vehicle pursuit without expert driver supervision. Specifically, during inference time our neural network operates in real-time and is capable of simultaneously predicting the commands for both longitudinal and lateral control from a single RGB camera and vehicle velocity streams.}

\item {We extensively test our method across a wide range of scenarios such as roundabouts, sharp turns and intersections using an autonomous driving simulator~\cite{Carla}. Our approach demonstrates higher tracking accuracy and robustness to variable conditions than the method inspired by~\cite{bojarski2016end}.}

\item {Training \& evaluation code along with videos showing qualitative results is available here: \href{https://changyaozhou.github.io/Autonomous-Vehicle-Pursuit/}{https://changyaozhou.github.io/Autonomous-Vehicle-Pursuit/}}.
\end{itemize}

\section{Related Work}
There have been multiple proposed solutions for the task of car-following, which can be roughly classified into two categories: those based on traditional techniques using control and optimization theory and those that deploy a neural network or a learning-based algorithm. 

\textbf{Control-based methods.}~\cite{7554530} involves a feedback controller to obtain desired velocity for steady platooning. Similarly, in~\cite{shi2021adaptive} a kinematic model, describing the motion of vehicles, is created. Then a speed controller is designed to adjust the vehicle's longitudinal and lateral speed. Meanwhile,~\cite{turri2017model} focuses on  energy-saving adaptive cruise control for eco-following. In our work, we leverage an MPC controller to collect training labels for our learning control method, which eliminates the necessity of manual annotation and recording of expert motion trajectories. ~\cite{660461} propose a two-layered control structure based on distance measurements and communication between the vehicles. In our case, we do not require a separate sensor for distance measurement. Rather, the distance and orientation are implicitly determined from an RGB image.

\textbf{Learning-based approaches.}~\cite{solomon2021follow} propose an end-to-end hierarchical deep neural network to follow a target pedestrian according to input images. In our case, vehicle pursuit poses an additional challenge due to the high and variable velocity of the target car. \cite{jahoda2020autonomous} designs a dual-task CNN, which simultaneously performs detection of the pursued object in 2D and semantic segmentation to determine drivable road regions. Based on the neural network estimates, a PID controller is utilized during inference time. In comparison, our approach neither requires semantic labels nor slow control optimization at inference time. We perform 3D object detection and MPC optimization offline and, thereby, ease the computational load at test time. \cite{bojarski2016end} utilizes a Convolutional Neural Network (CNN) for the prediction of steering control values given a three-camera setup, where each camera is forward-facing and laterally displaced. This work is the closest to our method. Nonetheless, our method does not require a third camera as we effectively render novel camera views using dense depth maps, which also makes our method more robust to noisy disturbances. Several recent works focus on the driving task via imitation learning by following a pre-defined route as in TransFuser~\cite{prakash2021multi} or relying on a high-level control planner policy~\cite{chen2022learning} that infers motion of all nearby traffic participants in a viewpoint-invariant manner. The ego-vehicle in our work, in contrast, follows a dynamic trajectory, which is unknown a-priori, and reactively adapts to the target car behavior. Moreover, our control method generalizes well to the viewpoint changes due to effective data augmentation. Furthermore, \cite{prakash2021multi} and its extension \cite{Chitta2022PAMI} deploy an additional LiDAR sensor along with the desired goal position to predict the future waypoints of the vehicle at inference time. In our work, we only use RGB images to determine the position and orientation of the goal (target vehicle) at inference and do not require extra LiDAR sensor to be present in the vehicle setup. ~\cite{nndriving2023} proposes a image-based learning method for lateral control. Our work in contrast presents a solution for both lateral and longitudinal control.

\section{Our Approach}
\label{sec:proposed approach}
In the next sections we provide a detailed overview of our proposed method for the car-following task as demonstrated in Fig.~\ref{overview}. Specifically, we describe three main steps of our pipeline: 1) label generation using Model Predictive Controller (MPC) and a 3D object detector (Sec. \ref{sec:label_generation}); 2) data augmentation via image rendering from novel views (Sec. \ref{subsec:data_augmentation}); 3) two-stage neural network training to predict control values for the ego vehicle such as throttle and steering angle to successfully follow the target car (Sec. \ref{subsec:nn}).

\subsection{Label Generation using MPC}
\label{sec:label_generation}
Ground truth data generation is carried out in a two-stage manner. Firstly, we leverage a 3D object detector to localize the target vehicle given sensor data from the ego-vehicle (Sec. \ref{subsec:3ddet}). Secondly, the obtained target pose together with velocity information is forwarded to the designed MPC algorithm in order to obtain optimal control values for the ego vehicle (Sec. \ref{subsec:mpc}). The label generation for car following can be seen as an alternative to an ``expert driver", which also allows richer and more diverse supervision~\cite{chen2020learning}. 

\subsubsection{3D Object Detection}
\label{subsec:3ddet}
In recent years, data-driven approaches have shown great performance for the task of 3D object localization in autonomous driving scenarios. Numerous camera-based \cite{li2019stereorcnn, qi2018frustum} and LiDAR-based methods \cite{lang2019pointpillars, Shi_2019_CVPR} have been proposed and extensively evaluated on various driving benchmarks. In this work, we leverage an off-the-shelf LiDAR-based 3D object detector and obtain 3D position $\boldsymbol{x} \in {\rm I\!R^3}$ and yaw angle $\theta$ of the target vehicle. In case of false positives, we prune the detections based on a simple, yet effective heuristic based on the distance to the ego-vehicle.

\subsubsection{Model Predictive Controller}
\label{subsec:mpc}
Model Predictive Control (MPC) is a method that is used to control a process while satisfying a set of constraints~\cite{camacho2013mpc}. It takes a prediction horizon instead of a single time step into account and aims to get an optimal control result by minimizing the cost function within the prediction horizon. 
In the following, we describe the cost function and dynamic model designed for our vehicle-pursuit scenario.

\textbf{Cost Function.} We consider several factors when designing the cost function for optimization. For brevity, the variables associated with the ego and target (reference) vehicle are denoted with the ${ego}$ and ${ref}$ subscripts respectively. 

To follow the target vehicle, the ego-vehicle must maintain a safe distance between the two vehicles to avoid a collision especially when a sudden application of brakes is applied by the target car. The safe distance $d_\text{safe}$ is formulated in proportion to the velocity of the target vehicle $d_\text{safe} = L \cdot (1 + w_1 \cdot v_{ref})$,
where $L$ should be at least the length of the vehicle and  $w_1 = 0.2$ is a scaling constant that is empirically chosen in our experiments.
Then the ideal distance between two vehicles along the x- and y-axis, namely $D_x$ and $D_y$, are computed respectively as 
\begin{align}
    D_x & = w_2 \cdot d_\text{safe} \cdot (\cos{\theta_{ego}} + \cos{\theta_{ref}})\\
    D_y & = w_2 \cdot d_\text{safe} \cdot (\sin{\theta_{ego}} + \sin{\theta_{ref}}),
\end{align}
where $\theta_{ego}$, $\theta_{ref}$ are yaw angles of the corresponding vehicles, $w_2 = 0.5$ is a scaling constant.

The \textit{positional cost term} to be minimized during the optimization is $cost_x = \left| (x_{ref} - D_x) - x_{ego}\right|$ and $cost_y = \left| (y_{ref} - D_y) - y_{ego}\right|$ computed separately for x-and y-axis based on the ideal distances $D_x$ and $D_y$ provided above. Moreover, the yaw angle of both vehicles should be as close as possible. Thus, we introduce an \textit{angle cost term} as $cost_{\theta} = (\theta_{ego} - \theta_{ref})^2$. Meanwhile, we assure that the estimated yaw angle $\theta_{ego}$ is within the range $[-\pi, \pi]$. The last component, \textit{velocity cost term} of the cost function computed as $cost_v = w_3 \cdot \left|v_{ref} - v_{ego}\right|$ is based on the velocities of both vehicles,
where a scaling constant $w_3 = 2$ is chosen empirically. The final optimization objective is defined as a sum of the aforementioned terms, namely the positional, angle and velocity cost terms.

In addition, we enforce a reasonable range of the lateral and longitudinal control values during optimization. Specifically, the range of throttle is [0, 1] and the range of steering angle is [-1, 1]. This range corresponds to the steering limits of the CARLA platform.

\textbf{Vehicle Model.} The model is used to update the state of the vehicle within the prediction horizon given the control values estimated by the optimization method.

Given the current state of the vehicle $x_{t}$, $y_{t}$, $v_{t}$ and $\theta_{t}$ we therefore obtain its next state at time $t + 1$ as 

\begin{equation}
\begin{aligned}[c]
x_{t+1} & = x_t + v_t \cdot \cos{\theta_t} \cdot dt \\
y_{t+1} & = y_t + v_t \cdot \sin{\theta_t} \cdot dt \\
\end{aligned}
\hspace{0.5cm}
\begin{aligned}[c]
v_{t+1} &= v_{t} + a_{t} \cdot dt - w_3 \cdot v_{t} \\
\theta_{t+1} & = \theta_t + (v_t \cdot dt \cdot \Phi ) / l, \\
\end{aligned}
\end{equation}
where $l$ is the length of vehicle and $w_3 = 0.05$. $a_{t}$ and $\Phi$ are proportional to throttle and steering angle respectively, which are optimized by the MPC.

With the updated states of the target and ego vehicles, the process repeats with a new iteration of cost function evaluation and optimal control values prediction. Despite the constant target velocity assumption in the optimization horizon, our full pipeline is robust to imperfect ground truth MPC labels and can handle sudden velocity changes of the target vehicle well.

\subsection{Data Augmentation}
\label{subsec:data_augmentation}
Small inaccuracies in the estimated lateral and longitudinal control values can cause variation in the perceived sensory data. Accumulation of such errors can even make the ego-vehicle lose track of the target car and crash. Hence, we propose a data augmentation approach, which allows the network to recover from divergences from the route being followed, thus generalizing to unseen scenarios and disturbances. Specifically, we include off-trajectory samples into consideration. While the on-trajectory samples are collected when the ego vehicle follows the target vehicle perfectly, the off-trajectory samples should be generated by sampling positions around the ideal trajectory. The off-trajectory positions of the ego-vehicle are computed by applying uniform offsets within some ranges to the on-trajectory position including longitudinal, lateral, and rotational offsets.

\subsubsection{Augmentation Steps}

To generate images from the view of the ego-vehicle, which is located at an off-trajectory position, image rendering can be used. For this, we only utilize on-trajectory RGB images and the corresponding dense depth maps. For depth map, we propose to utilize existing stereo-based depth estimation methods~\cite{div2020wstereo, hirschmuller2007stereo}.

Our data augmentation approach comprises the following steps. Firstly, we generate a local pointcloud by backprojecting every image pixel to the 3D camera coordinate frame given a dense depth map. Then we select an offset and transform the pointcloud from the on-trajectory ego-vehicle coordinate frame to the off-trajectory coordinate frame by applying a rigid body transformation $\boldsymbol{T}_{ego}^{ego'} \in SE(3)$. Thirdly, the points are projected to the image plane by applying the inverse operation to backprojection. This way, we warp image pixels from one view onto another.

Although we introduce off-trajectory samples, we do not need to perform 3D object detection from Sec.~\ref{subsec:3ddet} for every novel view. Instead, we can directly compute the relative transformation between off-trajectory ego vehicle coordinate and target vehicle coordinate according to Eq.~\eqref{eq:coord_transform} and use it as a training label for our neural network, where $ego$ and $ego'$ represent on- and off-trajectory ego vehicle respectively, $\boldsymbol{T}_{ref}^{ego'} \in SE(3)$ represents a rigid transformation between the coordinate frame of the off-trajectory ego-vehicle and the target vehicle.
\begin{equation}
\label{eq:coord_transform}
\boldsymbol{T}_{ref}^{ego'} = \boldsymbol{T}_{ego}^{ego'}\cdot \boldsymbol{T}_{ref}^{ego}
\end{equation}
$\boldsymbol{T}_{target}^{ego}$ can be extracted from the 3D object detector (Sec.~\ref{subsec:3ddet}), while $\boldsymbol{T}_{ego}^{ego'}$ is computed according to the longitudinal, lateral, and rotational offsets to the on-trajectory ego-vehicle pose.

\subsubsection{Sampling distances}
    
In this section we explain the sampling distances at which additional off-trajectory camera views are generated, which makes our model robust to noisy control values. We synthesize novel views at positions sampled from a uniform distribution every 0.2 meters in the lateral x-direction for up-to 3 meters on either side of the on-trajectory data. In addition, we add small random noise to sampled points. Specifically, sampling noisy offsets $x_{off}$ can be formulated as $x_{off} =  k \cdot 0.2 - \frac{N}{10} + \epsilon$, where $k \in \{0, 1 \ldots N\}$ and $\epsilon \sim \mathcal U(-0.05, 0.05)$. Meanwhile, for each of these lateral offset positions we sample longitudinal ($y_{off} \sim \mathcal N(0, 0.66)$) and rotational ($\theta_{off} \sim \mathcal N(0, 0.05)$) offsets. This way, 30 additional off-trajectory samples are collected at each frame. Given the offset values, a transformation matrix $\boldsymbol{T}^{ego'}_{ego}$ between the off-trajectory vehicle and on-trajectory vehicle is computed.

Then the relative transformation between the coordinate frame of the off-trajectory ego-vehicle and the target vehicle $\boldsymbol{T}^{ego'}_{ref}$ can be computed, as defined in Eq.~\eqref{eq:coord_transform}.

\begin{figure*}[h]
    \centering
    \includegraphics[width=1\textwidth]{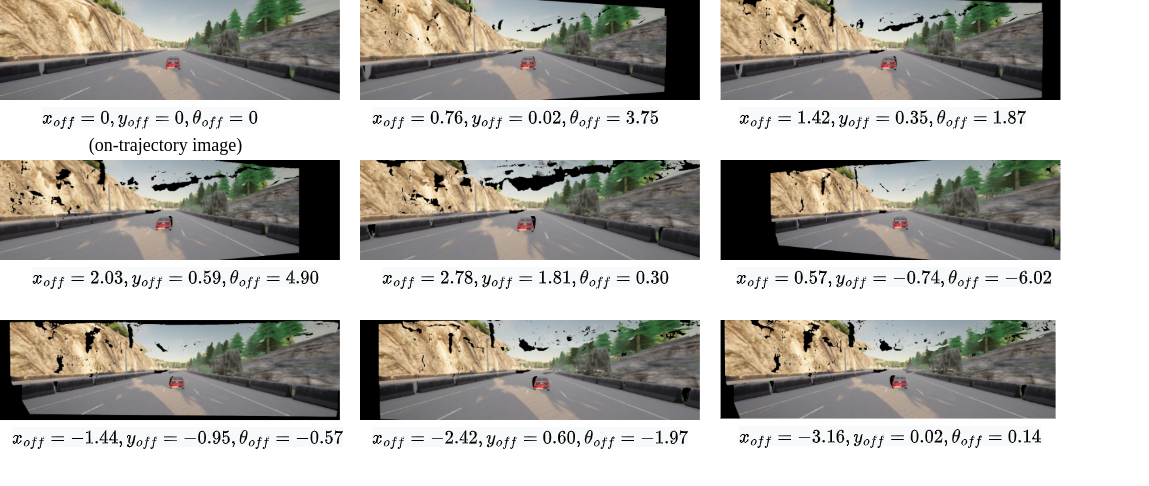}
    \caption{\label{render_example}The figures show some examples of rendered images with varying lateral ($x_{off}$), longitudinal ($y_{off}$) and rotational ($\theta_{off}$) offset from the original camera position. Note that the values of lateral and longitudinal offsets are in meters and the values of rotational offsets are in degrees. } 
    \end{figure*}

As we render novel views with lateral displacement given a single depth map, void regions are inevitable to obtain due to limited field of view, especially at image borders. We postprocess our rendered images by applying central crop and resizing before feeding them to the network. Nonetheless, our method for car following is robust against the remaining void regions caused by occlusion or discontinuities in the predicted depth map. \cite{8968451} showed that the sensorimotor control model focuses on high-level features such as lane markings, and pavement for immediate decision making. Thus, our processed rendered images still contain relevant information for the model to predict correct control commands. Fig. \ref{render_example} shows some visual examples of rendered images with various lateral, longitudinal and rotational offsets. More examples can be seen on our Project Page: \href{https://changyaozhou.github.io/Autonomous-Vehicle-Pursuit/}{https://changyaozhou.github.io/Autonomous-Vehicle-Pursuit/}.

It is worth noting that attempting to collect such novel views in real life by physically driving the car off-course may be too dangerous as the vehicle may invade lanes, sidewalks etc. thereby risking the safety of other road participants.

\subsection{Neural Networks}
\label{subsec:nn}

After the data augmentation step discussed above, we can now train the models. As shown in Fig.~\ref{overview} (right), the whole pipeline consists of two parts, that are trained separately: 

\emph{1) A Convolutional Neural Network (CNN)} model takes an RGB image from the ego and velocities from both vehicles as input and estimates the relative transformation between the two vehicles. Here we use the AlexNet~\cite{krizhevsky2017imagenet} model pre-trained on ImageNet~\cite{deng2009imagenet} as backbone. After flattening, we add linear layers to do regression for relative transformation.

\emph{2) A Multi-layer Perceptron (MLP)} model takes the predicted relative transformation as well as velocities as input and predicts the longitudinal and lateral control values (throttle, steering angle) for the ego-vehicle. The MLP model consists of five successive units, each containing a linear layer with ReLU as the activation function, a batch normalization layer and a final extra linear layer.

We found that additionally providing velocity information of the two vehicles to the CNN model improves its performance. We believe this information facilitates providing scale information to the model for determining the metric transformation while given a single-camera RGB stream.

\subsubsection{Training CNN}
\label{subsubsec:CNN}

The inputs of the CNN model are an RGB image sequence captured by a monocular camera mounted on the ego vehicle and velocities for both vehicles detected by speed sensors. The training labels are relative transformations between two vehicles, which have been obtained from a 3D object detector (Sec.~\ref{subsec:3ddet}). A 6-layer MLP model is designed for training, combining ReLU activation and Batch normalization.

The CNN model is trained with both on-trajectory data from 9 trajectories, and off-trajectory data generated through image rendering. The collected dataset contains around 210k images, with 180k allocated for training and 30k for validation. The model is successively trained with three different learning rates for 100 epochs in total (1e-3 for 20 epochs, 1e-4 for 60 epochs, and 1e-6 for 20 epochs).

\subsubsection{Training MLP}
\label{subsubsec:MLP}
As shown in Fig.~\ref{overview} relative transformations predicted by the CNN model along with the velocities of both vehicles are fed into an MLP model, which is used to predict the control values, namely the throttle and steering angle. The corresponding control values generated by the MPC controller (Sec. \ref{subsec:mpc}) in each frame are utilized as ground truth labels for training the MLP model. 

The MLP model is only trained with on-track data from 6 trajectories. The size of the dataset is around 30k samples, from which 24k samples are prescribed for training and 6k for validation. We train the model for 350 epochs with the learning rate 1e-4 and 1e-6 successively. 

After training, given the relative transformation predicted by the former CNN model and the velocities of both vehicles, the MLP model can predict reasonable control values, leading the ego vehicle to follow the target vehicle stably.

Note that both the CNN and MLP are trained using the Mean Squared Error (MSE) as the loss function.

\section{Experiments}
\label{sec:experiments}
\subsection{Experimental Setup}

Algorithms for sensorimotor control involve online interaction with the environment. Unfortunately, common autonomous driving benchmarks do not provide an option to conduct online interaction. Therefore, we use the CARLA driving simulator (version 0.9.11)~\cite{Carla} for our experiments. CARLA has been widely adapted for online evaluation in control algorithms such as~\cite{prakash2021multi,Chitta2022PAMI,jahoda2020autonomous} due to a wide range of sensors that can be attached to the ego-vehicle for both data collection and inference, as well as having maps with different terrains and routes. It also furnishes precise ground truth labels for various tasks such as semantic segmentation, optical flow, dense depth, etc. Real-time vehicle data, such as global position, orientation, velocity etc. can also be extracted from the simulator. The ability of simulators to test vehicle pursuit on various scenarios is very pertinent for our experiments. Nonetheless, it is worth mentioning that driving simulators cannot fully reproduce the full range of interactions in the real world. Specifically, other vehicles and pedestrians operate in a controlled manner as they have been modeled to follow a certain behavior. Moreover, simulators cannot fully replicate the physical hardware as they rely on mathematical models and approximations. For instance, a simulator may not be able to model sensor degradation under different circumstances such as heavy rain in the case of LiDAR. Despite these disadvantages, we believe that simulators remain a valuable tool for our work as a cost-effective and flexible alternative to the hardware implementations.

\textbf{Data Collection.} Data were collected from 9 trajectories across Towns 03 and 04. The target vehicle is run on autopilot by the Traffic Manager. Meanwhile, the ego vehicle is controlled by an MPC algorithm to follow the target while maintaining an appropriate safety distance. LiDAR and RGB sensors are placed on the ego-vehicle. We use the approach from \cite{Shi_2019_CVPR} to determine the 3D position and orientation of the target vehicle using LiDAR data. This information is in turn used by the MPC to determine the appropriate throttle and steering angle values as labels for training the neural network. At data collection, the RGB images are of size 1200 x 352 which are, firstly, center-cropped to 600 x 350 to remove the void regions and then resized to 128 x 128 before being fed to the CNN. Note that at inference time, only the RGB camera is used. The only information that we assume is given is the speed of the target and ego-vehicle. This can easily be read from a car's speedometer measurements.

The overall performance is evaluated on 5 trajectories from Towns 03 and 04 that are to a large extent different from the routes collected during the training phase. Since we noticed their partial overlap with the training routes, we further test the generalization capability of our method by adding 5 trajectories from two completely unseen maps, i.e. Town 01 and 05. While we describe quantitative results in Sec.~\ref{subsec:comparisons}, we also provide videos that show qualitative performance of our model on different scenarios: \href{https://changyaozhou.github.io/Autonomous-Vehicle-Pursuit/}{https://changyaozhou.github.io/Autonomous-Vehicle-Pursuit/}.

\subsection{Metrics}
\label{subsec:metrics}
To quantitatively evaluate the performance of vehicle pursuit, we adapt the Infraction Count (IC) and Route Completion (RC) metrics from \cite{prakash2021multi}. Higher IC and RC values indicate better system performance. Specifically for our application of vehicle pursuit, we introduce two additional metrics, namely Mean Translation Error (MTE) and Control Difference (CD). MTE captures how accurately the ego-vehicle can replicate trajectories of the target vehicle. CD represents the accuracy of the model in estimating the control values at each matching point. In order to compute Mean Translational Error values, we first find corresponding matching points in the ego and target trajectories using bipartite matching with the Hungarian algorithm \cite{kuhn1955hungarian}. Formally, we define MTE as
\begin{equation}
\label{MTE}
    MTE = \frac{1}{N}\sum_{i=1}^{N}||T_i||^2_2,
\end{equation}
where $N$ is the number of matching points and $||T_i||_2$ is the Euclidean norm of relative translational error at point $i$. Control Difference can be computed with 
\begin{equation}
\label{CD}
    CD = \frac{1}{N}\sum_{i=1}^{N} \sqrt{(a_{ego_i} - a_{ref_i})^2 + (\delta_{ego_i} - \delta_{ref_i})^2},
\end{equation}
where $a_{ego_i}$ and $a_{ref_i}$ are predicted and target vehicle throttle, $\delta_{ego_i}$ and $\delta_{ref_i}$ are predicted and target vehicle steering angle at point $i$ respectively.

\subsection{Quantitative Results}
\label{subsec:comparisons}
\begin{table}[htbp]
\vspace*{-5mm}
    \centering
    \caption{\label{table:compare_baselines}Performance of different methods.}
    \scalebox{0.9}{
    \begin{threeparttable}
    \begin{tabular}{cccccc} 
        \toprule
        \makecell[c]{Models} & RC $\uparrow$ & IC $\uparrow$ &  MTE $\downarrow$ & CD $\downarrow$\\
        \midrule
        Baseline & 36.87 & 0.01 & 2.90 & 0.29  \\
        Three-camera \cite{bojarski2016end} & 40.09 & 0.11 & 2.15 & 0.48 \\
        Our approach (SS Depth)\tnote{a} & 90.72 & \textbf{0.49} & \textbf{0.61} & \textbf{0.21} \\
        Our approach (Stereo Depth)& \textbf{90.88} & 0.41 & 0.80 & 0.24 \\
        \midrule
        GT Depth+ GT Transformation (Oracle)\tnote{b}& 92.90 & 0.42 & 0.57 & 0.17 \\
        \bottomrule
    \end{tabular}
    \begin{tablenotes}
    \item[a] SS = Self-Supervised;
    \item[b] GT = Ground Truth;
    \end{tablenotes}
    \end{threeparttable}
    }
    
    \vspace{-0.3cm}
\end{table}

Using metrics introduced in Sec.~\ref{subsec:metrics}, we evaluate our approach against different approaches, described in detail in the following subsections. As can be seen in Table \ref{table:compare_baselines}, our final approach outperforms all other methods and achieves performance comparable to that of the oracle. This is despite the oracle being trained with ground truth data. Table~\ref{table:compare_baselines} also demonstrates that our pipeline is not dependent on one particular approach for depth prediction.  It works equally well for both the self-supervised neural depth estimation network~\cite{div2020wstereo} and the stereo based classical approach~\cite{hirschmuller2007stereo}. Moreover, in case of the latter approach, our network is capable of handling inaccurate or noisy depth estimates. We now discuss details of the various model configurations described in Table \ref{table:compare_baselines}.

\subsubsection{Baseline}
This is the simplest approach with the model being trained on the on-trajectory images only. As shown in in Table \ref{table:compare_baselines}, this method has the lowest performance. This is because, at inference time, the model may cause the ego-vehicle to diverge from its driving lane, resulting in the vehicle being exposed to laterally displaced images. Since the model is only trained on on-trajectory images, it would not be able to correct off-trajectory motion and bring itself back on track. Accumulated displacements will cause the ego-vehicle to diverge and eventually lose track or crash.

\subsubsection{Three-camera Method}
The performance of the baseline model can be improved by collecting off-trajectory images by mounting multiple cameras at different positions on the ego vehicle. In this model configuration, we adapt the strategy of~\cite{bojarski2016end} by mounting three cameras on the ego vehicle, where each camera is placed with a 0.5-meter lateral displacement. The middle camera would capture on-trajectory images and the other two would capture off-trajectory images. Thus, this model is trained with three times more samples than the baseline model. 

Although the performance of this model in terms of RC and IC is better than the baseline model, the improvement is only marginal. The reason is that the images from the additional cameras are only at an offset of 0.5 m from the on-trajectory position. Therefore, if the ego-vehicle diverges beyond this limit at inference time, it will have difficulty recovering. A solution for further improvement could be to place the additional cameras at further distances. But this is difficult, as the distance between the cameras is constrained by the width/dimensions of the car. Moreover, adding more cameras could lead to higher costs and synchronisation challenges. 

\subsubsection{Our approach (Self-Supervised / Stereo Depth)}\label{subsec:our_experiment}
We demonstrate the performance of our method described in Sec. \ref{sec:proposed approach} with different depth estimation methods, which affect the quality of rendered off-trajectory images. In particular, we consider depth maps from a self-supervised network~\cite{div2020wstereo} and from a classical Semi-Global Block Matching (SGBM) method~\cite{hirschmuller2007stereo}. To determine relative transformation of the target vehicle with respect to the ego-vehicle we use the pre-trained model of \cite{Shi_2019_CVPR}.

As shown in Table \ref{table:compare_baselines}, the performance of our model is far superior to both the baseline and three-camera models. Meanwhile, its performance is comparable to the oracle. Due to image rendering, we are able to provide our network with significantly higher number of off-trajectory views than the baseline or three-camera methods. At the same time, our solution does not require additional sensors or the car to physically move off-trajectory. This can be viewed as the strength of our method and attributed to high performance against other approaches.

\subsubsection{Ground Truth Depth + Ground Truth Transformation (Oracle)}
\label{sssubsec:oracle}
Here, the ground truth depth and ground truth relative transformation of the target vehicle collected from the CARLA simulator are used to train the model. Since the method fully relies on the ground truth measurements, it is defined as the Oracle. Nonetheless, based on Table~\ref{table:compare_baselines}, the performance of this model is marginally better than our method, despite the fact that our approach is trained without ground truth labels.

\subsection{Additional Experiments}

We conduct an ablation study and further experiments to elaborate and assess different aspects of our method.

\subsubsection{Impact of Involving Ground Truth (GT) Data}

As mentioned in Sec. \ref{sssubsec:oracle}, we view the model trained with GT depth map and GT transformation as the Oracle. However, the GT data (collected in the CARLA simulator) might not always be accessible. Therefore, we perform additional experiments and investigate the extent to which the performance of our model is affected if the GT measurements are not available during training. Two variations for ``GT Depth" and ``GT transformation" models in Table \ref{table:compare_baselines} are adopted.

Specifically, ``GT Depth + 3D Object Detector" utilizes ground truth dense depth maps from the CARLA simulator for image rendering. Meanwhile ``SS Depth + GT Transformation" uses ground truth position and orientation of the target vehicle from the CARLA simulator. This information is then directly used for the control label generation using MPC as described in Sec.~\ref{subsec:mpc}.

The first 4 rows of Table \ref{table:ablation_gt} show the evaluation results of 4 models trained with and without ground truth depth map and ground truth transformation, where the performance is compared to the Oracle. ``GT Depth + 3D Object Detector" model demonstrates that using 3D object detections instead of the ground truth transformation labels for training the CNN module does not negatively influence the performance. Moreover, the performance of the ``SS Depth + GT Transformation" model also indicates that the depth map predicted by the self-supervised model is sufficiently accurate and compares well with the GT depth map. Thus, the images rendered using self-supervised depth maps would look similar to those rendered using GT depth.
\begin{table}[htbp]
    \centering{
    \caption{\label{table:ablation_gt}Evaluation results of methods with or without ground truth data.} 
    \begin{threeparttable}
    \scalebox{1.0}{
    \begin{tabular}{cccccc} 
        \toprule
        \makecell[c]{Models} & RC $\uparrow$ & IC $\uparrow$ & MTE $\downarrow$ & CD $\downarrow$\\
        \midrule
        GT Depth+GT transformation\tnote{a}& \textbf{92.90} & 0.42 & 0.57 & \textbf{0.17} \\
        GT Depth+3D Detector& 91.13 & 0.46 & 0.62 & 0.19 \\ 
        SS Depth+GT transformation& 90.42 & \textbf{0.51} & \textbf{0.54} & 0.19 \\
        \makecell{Our approach\\(SS Depth+3D Detector\tnote{b})} & 90.72 & 0.49 & 0.61 & 0.21 \\
        \midrule
       \makecell[c]{Our approach \\ (with Transformation labels)} & 100 & 0.19 & 0.85 & 0.10 \\
        \makecell[c]{Our approach \\ (with MPC controller)}& 91.30 & 0.13 & 0.78 & 0.13 \\
        \bottomrule
    \end{tabular}
    }
    \begin{tablenotes}
    \item[a] GT = Ground Truth;
    \item[b] SS = Self-Supervised;
    \end{tablenotes}
    \end{threeparttable}
    }

\end{table}   
In the last two rows of Table \ref{table:ablation_gt}, we investigate the performance of the transformation (CNN) and control (MLP) modules independently. To test for the MLP, we use the actual transformation labels obtained directly from the simulator as input, instead of the transformation labels predicted by the CNN. Likewise, to test for the CNN, we feed the predicted transformation output to an MPC controller, instead of the MLP.  The results show that the MLP behaves very well when provided with precise transformation labels. It reaches 100 RC score, implying that the gap in RC performance likely comes from the limitations of the CNN architecture which can possibly be reduced in the future by using a more powerful architecture. However, note that the IC and MTE metrics are worse off.    Similarly, replacing the MLP with MPC does not show any significant performance enhancement. This demonstrates that the MLP is already achieving performance at par with the MPC, given the noisy labels from the transformation module.

\subsubsection{Choice of Point Cloud Source for 3D Object Detection}
\label{sec:object detection experiment}
As described in Sec.~\ref{subsec:3ddet} our method utilises point clouds from the LiDAR sensor as the input to the object detection algorithm for determining the state of the target vehicle. MPC then utilizes this state to generate the training labels for the CNN network. 
In this experiment, we investigate the accuracy of 3D object detection using point cloud data from different sources other than LiDAR such as stereo cameras. For this, we utilize the same pre-trained 3D object detector~\cite{Shi_2019_CVPR} and test it on both LiDAR scans (Ours) and point clouds generated from stereo images. In particular, we consider two stereo approaches, namely SGBM~\cite{hirschmuller2007stereo} and CDN~\cite{div2020wstereo}. As it can be observed in Table \ref{tab:errors}, LiDAR scans are more accurate for the 3D object detection task when compared to the image-based depth prediction methods. Therefore, it validates our choice of LiDAR point cloud as inaccuracies in 3D detection would significantly deteriorate the control labels from the MPC for our training. It is however important to note that due to the sparsity of LiDAR point clouds, they cannot be used to synthesize images. Hence, image-based depth estimation methods are needed for image generation as elaborated in Section \ref{subsec:data_augmentation}

\begin{table}[htbp]
    \centering{  
    \caption{\label{tab:errors}Mean Squared Error of Object Detection with Different Point Cloud Sources} 
    \begin{threeparttable}
    \begin{tabular}{cccc} 
        \toprule
        \textbf{Point Cloud Source} & $MSE_x$\tnote{a} $\downarrow$ & $MSE_y$\tnote{b} $\downarrow$ & $MSE_{\theta}\tnote{c}$ $\downarrow$ \\
        \midrule
        SGBM\cite{hirschmuller2007stereo} & {1.50} & {3.00} & {0.24}\\
        CDN\cite{div2020wstereo} & {4.89} & {3.97} & {0.77}\\
        LiDAR (Ours) & \textbf{0.18} & \textbf{0.05} & \textbf{0.04}\\          
        \bottomrule
    \end{tabular}
    \begin{tablenotes} 
        \item[a]lateral MSE  
        \item[b]longitudinal MSE
        \item[c]rotational MSE
    \end{tablenotes}
    \end{threeparttable}
    }
\end{table}

\subsubsection{Performance on Different Levels of Perturbation}

We also tested the stability and robustness of our model against the baseline by applying external force with different intensity levels in random directions to the ego vehicle as perturbations. The external force can be according to $F_{perturb} = 2\cdot L \cdot m_{ego}$, where L represents the level of perturbation and $m_{ego}$ is the mass of the ego vehicle.

\begin{figure}[h!]
\centering
\includegraphics[width=0.45\textwidth]{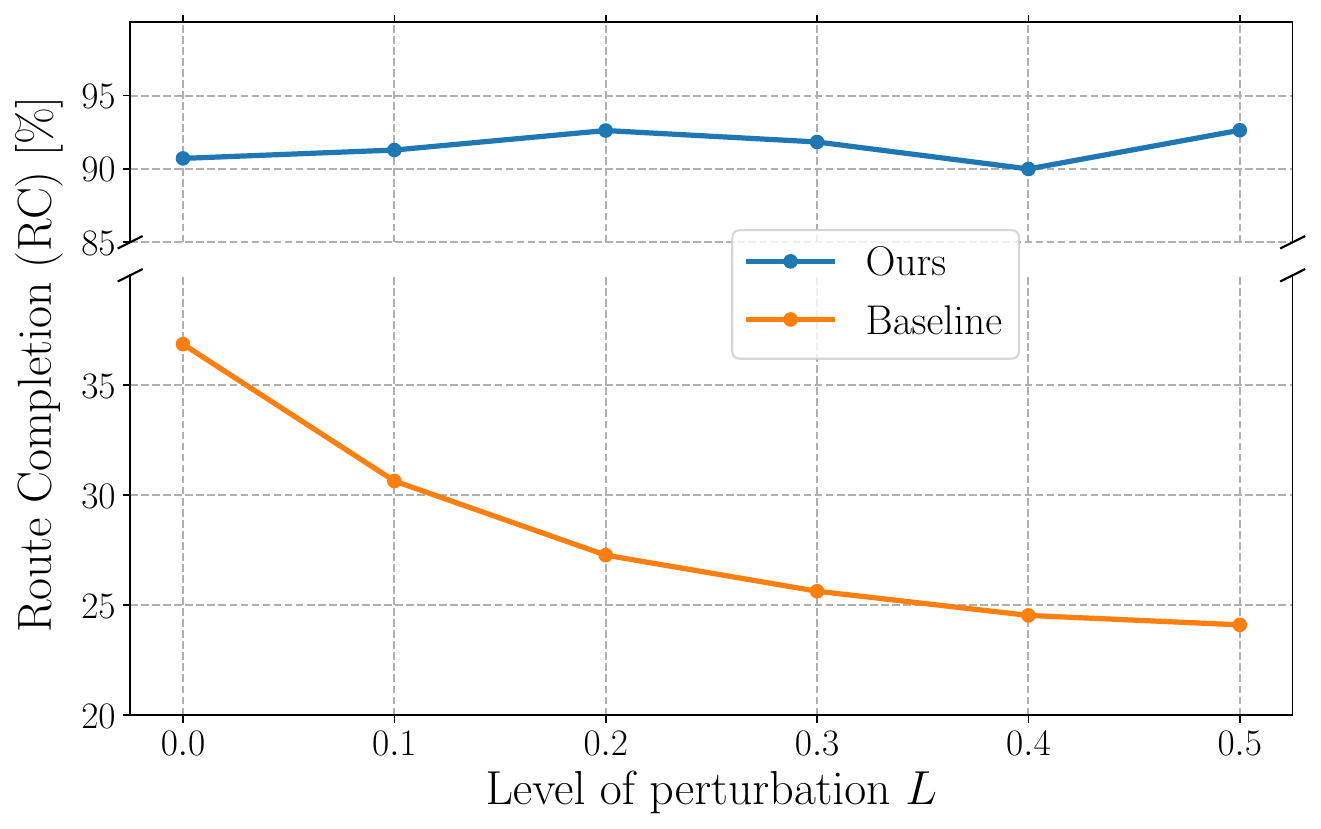}
\caption{\label{evaluation_impulse} Average Route Completion (RC) under different perturbation levels. Unlike the baseline, our model maintains high accuracy \& consistency with increasing perturbations.}
\vspace{-0.2cm}
\end{figure}

The external forces are added to the ego vehicle for three continuous frames every 40 frames. Applying such perturbation would cause the vehicle to diverge from its normal path. The subsequent 37 frames allow the model to recover from this divergence thereby leading the vehicle back to follow the target vehicle again. As shown in Fig.~\ref{evaluation_impulse}, the route completion of our method remains almost unchanged despite the increasing level of perturbation. However, the performance of the baseline model decreases dramatically. This experiment demonstrates that our model trained with rendered off-trajectory images is capable to deal with unexpected perturbations. 

\subsubsection{Performance in Different Target Velocity \& Acceleration ranges}

In our method, the target vehicle velocity \& acceleration are crucial variables for successful vehicle pursuit.  The ego-vehicle being controlled should closely match the target vehicle’s velocity \& acceleration. This means that the velocity of the ego-vehicle should rise (or drop) with the increase (or decrease) in the target vehicle’s velocity. Otherwise, the ego-vehicle would either not be able to keep up with the target or collide into it.  

\begin{figure}[h!]
\centering
\includegraphics[width=0.45\textwidth]{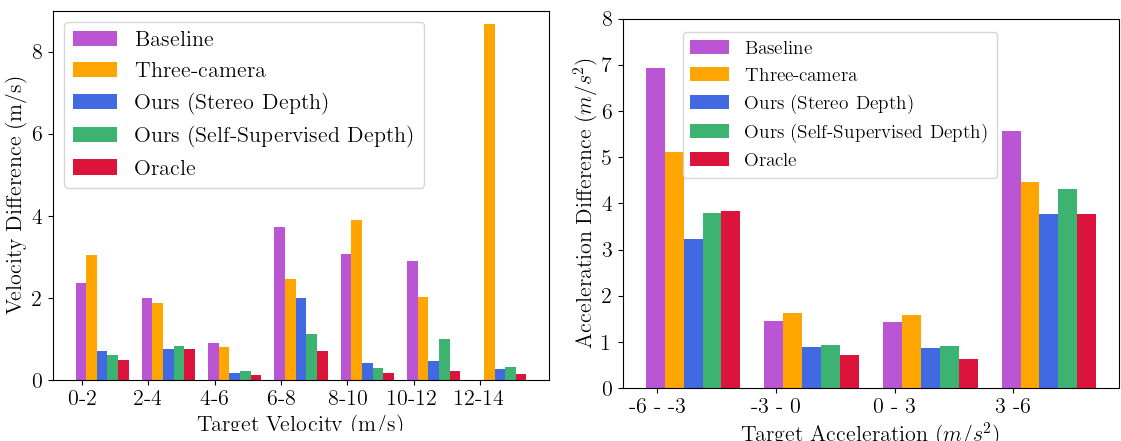}
\caption{\label{vel_acc_plot} The mean velocity (left) and acceleration (right) difference between target and ego vehicle for different ranges.}
\vspace*{-2mm}

\end{figure}

To assess the adaptability of all models to different target velocity \& acceleration ranges, we calculate the mean velocity/acceleration difference between the target and ego vehicles within different ranges. Hence, the locations of target and ego vehicles over each trajectory are first matched with each other by the closest distance, then the velocity/acceleration differences are computed. As seen in  Fig.~\ref{vel_acc_plot}, both our models with Self-Supervised and Stereo Depth show a lower velocity/acceleration  difference in comparison to the Baseline and Three camera model. In fact, they demonstrate performance comparable to that of the Oracle. 

\subsubsection{Computational Analysis}
We further test the runtime performance of our model during inference (online performance). When tested on a low-end GPU GeForce MX250 with 384 cores, our method achieves an inference speed of 42 frames per seconds (fps), which can be considered for a real-time application system. 
In contrast, running the object detection method and MPC optimization at inference time would lead to a slower runtime. In fact, the 3D object detector alone yields only 1.5 fps.

\section{Limitations}
\label{limitations}

We assume that we are only following one target vehicle which is seen at all times. If the target vehicle briefly moves beyond the field of view of the camera or is completely occluded by another vehicle then the model would struggle to follow the target even if it appears back later. An option to address this in future work is to integrate tracking based Vehicle Re-Identification techniques from works such as~\cite{KHAN201950}. For this, our architecture would need to be slightly modified to take in the template of the target vehicle too. Moreover, with this solution, a single target vehicle to be followed can be selected among the multiple vehicles in the scene.   Another limitation is that we assume no uncertainty or drop in the communication of velocity value between two vehicles. A solution to the noisy or missing target vehicle velocity reading is to apply approaches such as the Kalman filter~\cite{kalman1960}.
\section{Conclusions}
\label{sec:conclusion}
In this paper, we proposed a car-following framework for training a network without the need for supervised steering labels. The steering labels are implicitly determined from model predictive control. This is done in conjunction with applying a 3D object detector 
to extract the relative location and orientation of the target vehicle with respect to the ego-vehicle. 
Given a single RGB image and velocities of both vehicles, we train a two-stage network comprising of a CNN and an MLP to estimate optimal lateral and longitudinal control values for the ego-vehicle while it performs the task of following the target. Meanwhile, additional off-trajectory images are rendered and included in the training data to enhance the robustness of the model to inaccuracies in estimated control commands. We extensively test our method on the CARLA simulator and show the effectiveness of our pipeline. Additional experiments are performed to quantitatively verify robustness and computational efficiency of our method.

Although the emphasis of the work was on the car-following task only, we believe the proposed approach can be extended to a multi-vehicle platooning application. We believe our work can serve as an intermediate step from vehicle pursuit to a fully autonomous driving and inspire future work in the same direction.


\bibliographystyle{IEEEtran}
\balance
\bibliography{IEEEabrv,IEEEexample}
\end{document}